\documentclass[bioengineering,article,accept,pdftex,moreauthors]{Definitions/mdpi} 

\firstpage{1} 
\pubvolume{1}
\issuenum{1}
\articlenumber{0}
\pubyear{2023}
\copyrightyear{2023}
\externaleditor{Academic Editor: {Firstname Lastname}}

\datereceived{} 
\daterevised{}
\dateaccepted{} 
\datepublished{} 
\hreflink{https://doi.org/} 

\usepackage{graphicx}
\usepackage{booktabs}
\usepackage{siunitx}
\usepackage{amsmath}
\usepackage{pifont}
\usepackage{amssymb}
\usepackage{fontawesome}
\usepackage{colortbl}
 \usepackage[labelformat=simple]{subcaption}

\Title{{Prompt-Based} Tuning of Transformer Models for Multi-Center Medical Image Segmentation of Head and Neck Cancer}

\TitleCitation{Prompt-Based Tuning of Transformer Models for Multi-Center Medical Image Segmentation of Head and Neck Cancer}

\Author{{Numan Saeed} $^{1,}$*$^{,\dagger}$\orcidA{}, Muhammad Ridzuan $^{1, \dagger}$\orcidB{}, Roba {Al Majzoub} $^{2}$\orcidC{} and Mohammad Yaqub $^{2}$\orcidD{}}

\AuthorNames{Numan Saeed, Muhammad Ridzuan, Roba Al Majzoub and Mohammad Yaqub}

\AuthorCitation{Saeed, N.; Ridzuan, M.; Al Majzoub, R.; Yaqub, M.}

\address{%
$^{1}$ \quad Department of Machine Learning, Mohamed bin Zayed University of Artificial Intelligence, {Abu Dhabi 7909,} 
 United Arab Emirates; muhammad.ridzuan@mbzuai.ac.ae (M.R.)\\

$^{2}$ \quad Department of Computer Vision, Mohamed bin Zayed University of Artificial Intelligence, {Abu Dhabi 7909}, United Arab Emirates; roba.majzoub@mbzuai.ac.ae (R.A.M.); mohammad.yaqub@mbzuai.ac.ae (M.Y.)}

\corres{Correspondence: numan.saeed@mbzuai.ac.ae (N.S.)}

\firstnote{These authors contributed equally to this work.} 

\abstract{Medical image segmentation is a vital healthcare endeavor requiring precise and efficient models for appropriate diagnosis and treatment. Vision transformer (ViT)-based segmentation models have shown great performance in accomplishing this task. However, to build a powerful backbone, the self-attention block of ViT requires large-scale pre-training data. The present method of modifying pre-trained models entails updating all or some of the backbone parameters. This paper proposes a novel fine-tuning strategy for adapting a pretrained transformer-based segmentation model on data from a new medical center. This method introduces a small number of learnable parameters, termed prompts, into the input space (less than 1\% of model parameters) while keeping the rest of the model parameters frozen. Extensive studies employing data from new unseen medical centers show that the prompt-based fine-tuning of medical segmentation models provides excellent performance regarding the new-center data with a negligible drop regarding the old centers. Additionally, our strategy delivers great accuracy with minimum re-training on new-center data, significantly decreasing the computational and time costs of fine-tuning pre-trained models. Our source code will be made publicly available.}


\begin{document}



\section{Introduction}
Recently, several novel segmentation models have been proposed to assist in medical image analysis and understanding, leading to faster and more accurate treatment planning~\cite{alalwan2021efficient,valanarasu2021medical,zhou2021nnformer}. Many of these proposed models are increasingly transformer-based, demonstrating excellent performance on several medical datasets. Transformers are a class of neural network topologies distinguished chiefly by their heavy usage of the attention mechanism~\cite{vaswani2017attention}. In particular, Vision transformers (ViTs)~\cite{dosovitskiy2020image} have demonstrated their ability in 3D medical image segmentation~\cite{hatamizadeh2022unetr,YAN2023109432}. However, ViTs exhibit an intrinsic lack of image-specific inductive bias and scaling behavior; nonetheless, this lack is mitigated by utilizing large datasets and large model capacity. 


On the other hand, medical datasets are limited in size due to time-consuming and expensive expert annotations, which hinders the use of powerful transformer models with regard to their full capacity. A common approach to handle the limited data size in the medical domain is to use transfer learning~\cite{medical_transfer_learning}. Multiple studies exploited pretrained networks for different downstream tasks such as classification~\cite{transfer_1}, segmentation~\cite{Karimi2021-zw}, and progression~\cite{WARDI2021395}. This technique aims to reuse model weights or parameters of already trained ViTs on different but related tasks. More specifically, models are first pretrained on a different large dataset; the pretraining weights act as informed initializations of the model~\cite{simclr,moco,lin2022vision}. The pretrained model is then fine-tuned on the target dataset, yielding faster training and a more generalizable model.

\begin{figure}[H]
\begin{adjustwidth}{-\extralength}{0cm}
\centering
\includegraphics[width=18cm]{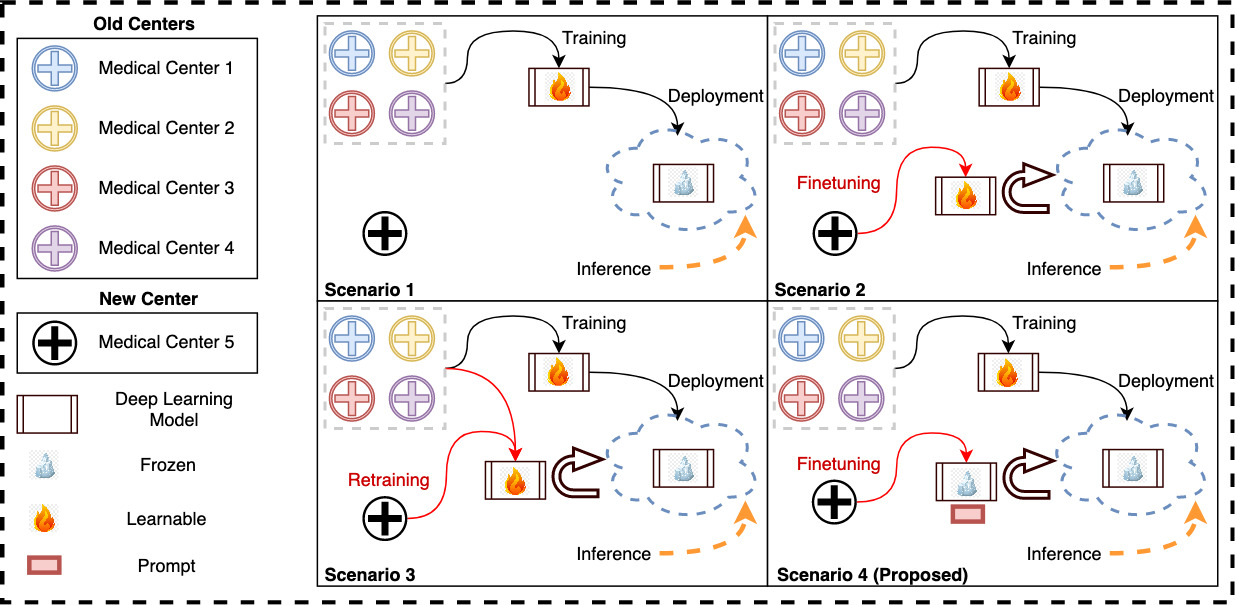}
\end{adjustwidth}
\caption{The four different scenarios of using the deployed deep learning model with the old and new medical centers' data. In (\textbf{Scenario 1}), 
 the new-center data is directly inferred through the deployed model trained on old-center data (no finetuning). In (\textbf{Scenario 2}), the model is fully or partially finetuned on the new-center data before being deployed for inference. In (\textbf{Scenario 3}), the model is retrained using both old- and new-center data before deployment. Our proposed method (\textbf{Scenario 4}) utilizes the data solely from the new center to finetune only the prompt while keeping the trained model frozen and then deploying it. \label{fig1}}
\end{figure}

However, the limited size of medical datasets is not the only challenge; medical datasets are sourced from different medical centers that use different machines and acquisition protocols, leading to further heterogeneity in the acquired data~\cite{glocker2019,ma2018}. As a result, a model trained on data obtained from specific medical centers might fail to perform well on data obtained from a new medical center, see Figure~\ref{fig1} (Scenario 1) . Conventionally, we can use the transfer learning technique for adapting the pretrained model to the new medical center data. One such effective adaptation strategy is partial/full fine-tuning, in which some/all of the parameters of the pretrained model are fine-tuned on the new center's data, see Figure~\ref{fig1} (Scenario 2). However, directly fine-tuning a pretrained transformer model on a new center's data can lead to overfitting (as we have mostly small-size datasets from any new center) and catastrophic forgetting (loss of knowledge learned from the previous centers)~\cite{barone2017,kumar2022}. Hence, this strategy requires storing and deploying a separate copy of the backbone parameters for every newly acquired medical center data. This strategy is costly and infeasible if the end solution is regularly deployed on new medical centers or the acquisition protocol and/or machines in an existing center change. Particularly, this infeasibility will be more prominent in transformer-based models as they are significantly larger than their convolutional neural network (CNN) counterparts. Another possibility is to re-train the model on samples from old and new centers data and re-deploy it upon inference, see Figure~\ref{fig1} (Scenario 3). This scenario is computationally expensive and infeasible due to the same pitfalls of Scenario 2.    

{In this work, inspired from ~\cite{lester2021power, li2021prefix, jia2022visual} we propose a prompt-based fine-tuning method of ViTs on new medical centers' data. It is important to note that previous studies have mainly focused on large language models ~\cite{lester2021power, li2021prefix} and natural images ~\cite{jia2022visual}. However, our research is centered around utilizing prompt-based fine-tuning to tackle medical image segmentation tasks. More specifically, we are looking at multi-class segmentation of cancer lesions with multi-center data.} Instead of altering or fine-tuning the pretrained transformer, we introduce center-specific learnable token parameters called prompts in the input space of the segmentation model. Only prompts and the output convolutional layer are learnable during the fine-tuning of the model on the new center's data. The rest of the entire pre-trained transformer model is frozen. Current deployment scenarios as well as our proposed approach (Scenario 4) are depicted in Figure~\ref{fig1}.

We show that this method can achieve high accuracy on new centers' data with a negligible loss regarding the accuracy of the old centers, in contrast to full or partial fine-tuning techniques, where the model accuracy comprises the old-center data. The main contributions of this work are as follows:
\begin{itemize}
    \item We propose a new {\textit{prompt}}
-based fine-tuning technique for the {\textit{transformer}}-based medical image segmentation models that reduces the fine-tuning time and the number of learnable parameters (less than 1\% of the model parameters) to be stored for the new medical center.
    \item The proposed method achieves equivalent accuracy for new-center data compared to the full fine-tuning technique while mostly preserving the accuracy for the old-center data that compromises full fine-tuning. 
    \item We showcase the efficacy of the proposed method on {\textit{multi-class}} segmentation of head and neck cancer tumors using {\textit{multi-channel}} computed tomography (CT) and positron emission tomography (PET) scans of patients obtained from {\textit{multi-center}} (seven centers) sources.
\end{itemize}

\section{Methodology}

\begin{figure}[H]
\begin{adjustwidth}{-\extralength}{0cm}
\centering
\includegraphics[width=18cm]{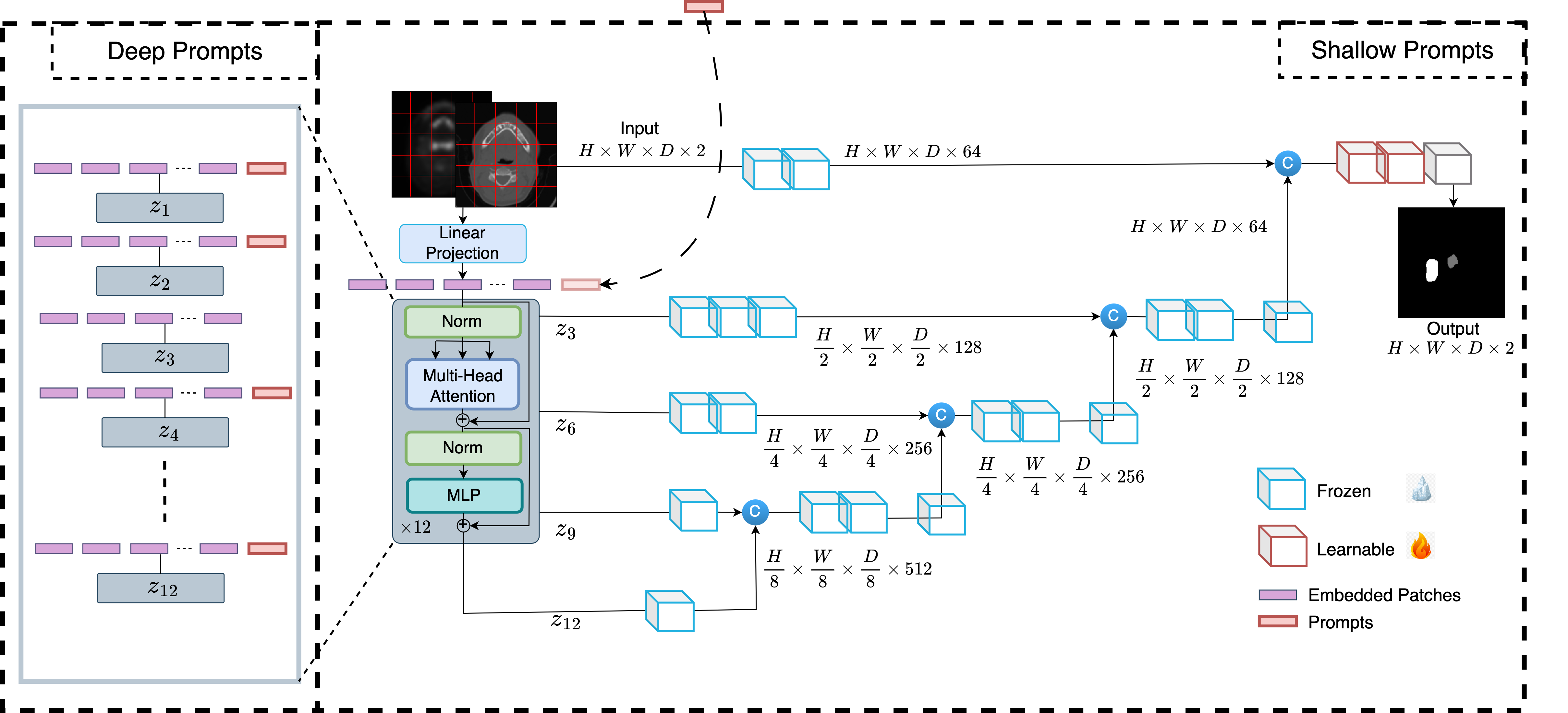}
\end{adjustwidth}
\caption{Overview of the proposed method. Learnable prompts are appended to the embedded tokens in the input space and passed through the transformer encoder but not the decoder during the fine-tuning. In deep prompt-based fine-tuning, the learnable prompts are replaced by new prompts after each transformer layer.\label{fig2}}
\end{figure}

Due to differences in how imaging is done, what equipment is used, and who the patients are, the quality and distribution of the data collected by different medical centers might be very different. This heterogeneity represents a barrier to developing precise and robust models that can generalize to new medical center data optimally. In this section, we describe a novel tuning technique, called prompt-based tuning, that overcomes the pitfalls of conventional fine-tuning techniques. In this section, we describe prompt-based tuning for adapting transformer-based medical image segmentation models. Prompt-based fine-tuning technique injects a small number of learnable parameters into the transformer's input space and keeps the backbone of the trained model frozen during the downstream training stage. The overall framework is presented in Figure \ref{fig2}. We demonstrate two variants of prompt-based tuning, shallow and deep, and compare their performance to the conventional fine-tuning methods such as partial and full fine-tuning. Below, we describe the two prompt-based tuning methods and highlight the differences between the two.

\subsection{Shallow Prompt Tuning}

In shallow prompt fine-tuning, a set of \textit{p} continuous prompts of dimension \textit{d} are introduced in the input space after the embedding layer. These prompts are concatenated with the token embeddings of the volumetric patches of an input image $x\in \mathbb{R}^{H\times W\times D\times C}$, where $H$, $W$, $D$, and $C$ are the height, width, depth, and channels of the 3D image, respectively. $K\times K\times K$ represents the dimensions of each patch, and $n=HWD/K^3$ is the number of patches extracted. The embedding layer projects these patches to a dimension $d$. The class token is dropped from the ViT~\cite{dosovitskiy2020image} as the experiments are for a segmentation task. The resulting concatenated prompts and embeddings are fed to a transformer encoder consisting of $L$ layers, following the same pipeline as the original ViT~\cite{dosovitskiy2020image}, with normalization, multi-head self-attention (MSA), and multi-layer perceptron. The decoder only uses image patch embeddings as inputs, and prompt embeddings are discarded. The shallow prompt-based fine-tuning is formulated as: 
\begin{gather}
    \qquad \quad x_0 = Embedding(x) \qquad\qquad\qquad\quad  x_0\in \mathbb{R}^{n\times d}\\
    [{U_1}, x_1] = Encoder_1([P,x_0]) \quad\qquad\qquad \quad P\in \mathbb{R}^{p\times d}\\
    [{U_i}, x_i] = Encoder_i([{U_{i-1}},x_{i-1}])  \qquad\qquad i= 2,...,L \\
    Y_{seg} = Decoder(ConvTrans3D(x_i)) \quad\qquad i = 3,6,9,12
\end{gather}
where $P$ is the prompt matrix and $ConvTrans3D$ refers to 3D transpose convolution.

\subsection{Deep Prompt Tuning}
In deep prompt fine-tuning, the prompts can be introduced at the input space of each transformer layer or subset of layers. In our implementation, we add the deep prompts after each skip connection layer:
\begin{equation}
   \label{dice_loss}
   [\_, x_i] = Encoder_i([P_{i-1},x_{i-1}])  \qquad\qquad i= 1,...,L 
\end{equation}

\vspace{-6pt}

\section{Experiments}
We use the state-of-the-art transformer-based segmentation models, UNETR~\cite{hatamizadeh2022unetr} and Swin-UNETR~\cite{hatamizadeh2022swin}. In addition, we compare the two variants of the proposed method to partial and full fine-tuning, two prevalent transfer learning protocols used in medical imaging. 

\subsection{Dataset}

The dataset used in this work is multi-center, multi-class, and multi-modal. This dataset comprises head and neck cancer patient scans collected from seven centers. The data consist of CT and PET scans, as well as electronic health records (EHR) of each patient. The PET volume is registered with the CT volume to a common origin, although they each have varying sizes and resolutions. The CT sizes range from (128, 128, 67) to (512, 512, 736), while the PET sizes range from (128, 128, 66) to (256, 256, 543) voxels. The CT resolutions range from (0.488, 0.488, 1.00) to (2.73, 2.73, 2.80), while the PET resolutions range from (2.73, 2.73, 2.00) to (5.47, 5.47, 5.00) mm in the $x$, $y$, and $z$ directions. Some scans are of the head and neck regions, while others contain the full body of the patients. 
\begin{figure}[H]
\captionsetup[subfigure]{justification=centering}
\begin{minipage}{0.18\textwidth}
\begin{subfigure}{\textwidth}
    \includegraphics[width=\textwidth]{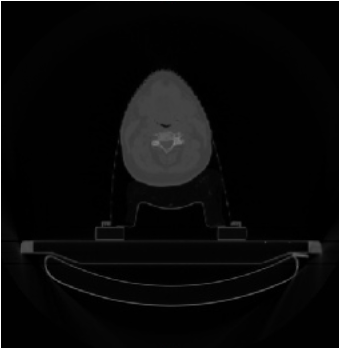}
    \subcaption{\textbf{}}
\end{subfigure}
\end{minipage}
\begin{minipage}{0.18\textwidth}
\begin{subfigure}{\textwidth}
    \includegraphics[width=\textwidth]{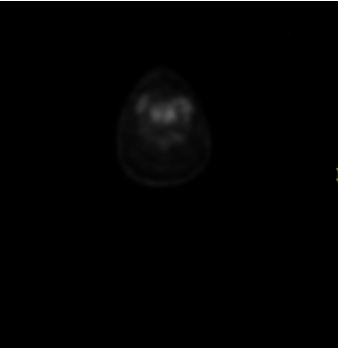}
    \subcaption{\textbf{}}
\end{subfigure}
\end{minipage}
\begin{minipage}{0.18\textwidth}
\begin{subfigure}{\textwidth}
    \includegraphics[width=\textwidth]{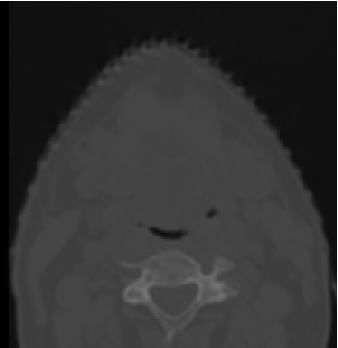}
    \subcaption{\textbf{}}
\end{subfigure}
\end{minipage}
\begin{minipage}{0.18\textwidth}
\begin{subfigure}{\textwidth}
    \includegraphics[width=\textwidth]{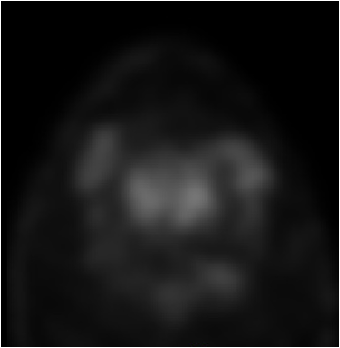}
    \subcaption{\textbf{}}
\end{subfigure}
\end{minipage}
\begin{minipage}{0.18\textwidth}
\begin{subfigure}{\textwidth}
    \includegraphics[width=\textwidth]{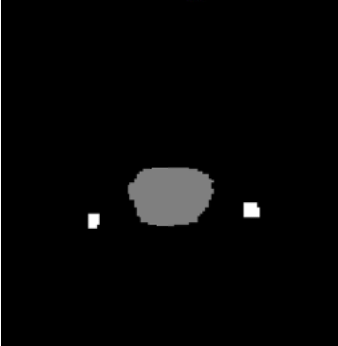}
    \subcaption{\textbf{}}
\end{subfigure}
\end{minipage}
\caption{A sample 
 of images from the dataset~\cite{oreiller2022head}. (\textbf{a},\textbf{b}) depict the original CT and PET scans, respectively. (\textbf{c},\textbf{d}) show the cropped CT and PET scans, and (\textbf{e}) shows the cropped ground truth mask.
}
\label{data_sample}
\end{figure}

The PET/CT scans are in the NIFTI format. They have been resampled to $1\times1\times1$ mm$^{3}$ isotropic resolution and cropped to a dimension of $176\times176\times176$  around the primary tumor and lymph nodes. The {CT HU value} is clipped to a range of $-$200 to 200, while the PET is clipped to a maximum of 5 standard uptake values (SUV).

The dataset contains segmentation masks for each patient, including the ground truth of primary gross tumor volumes (GTVp), nodal gross tumor volumes (GTVn), and other clinical information. The annotations were made by medical professionals at the respective centers and are provided with the dataset. The dataset is publicly available on the MICCAI 2022 HEad and neCK TumOR (HECKTOR) challenge website~\cite{oreiller2022head}. The complete dataset consists of 524 samples. The detailed distribution of the dataset across different centers is listed in Table \ref{tab1} along with the {type of scanner} used to acquire the scans. 


\begin{table}[H] 
\caption{Dataset origin and distribution.\label{tab1}}
\newcolumntype{C}{>{\centering\arraybackslash}X}
\begin{tabularx}{\textwidth}{CCCC}
\toprule
\textbf{Center}	& \textbf{City, Country}	& \textbf{PET/CT scanner} & \textbf{Number of samples} 
\\
\midrule
HGJ & Montreal, Canada &  Discovery ST, GE Healthcare & 55\\
CHUS & Sherbrooke, Canada & GeminiGXL 16, Philips & 72\\
HMR & Montreal, Canada & Discovery STE, GE Healthcare & 18\\
CHUM & Montreal, Canada & Discovery STE, GE Healthcare & 56\\
CHUV & Vaud, Switzerland & Discovery D690 TOF, GE Healthcare & 53\\
CHUP & Poitiers, France & Biograph mCT 40 ToF, Siemens & 72\\
MDA & Texas, USA & Discovery HR, RX, ST, and STE {(GE Healthcare)} 
 & 197\\
\bottomrule
\end{tabularx}
\end{table}

\subsection{Experimental Setup}

\begin{figure}[H]
\includegraphics[width=\textwidth]{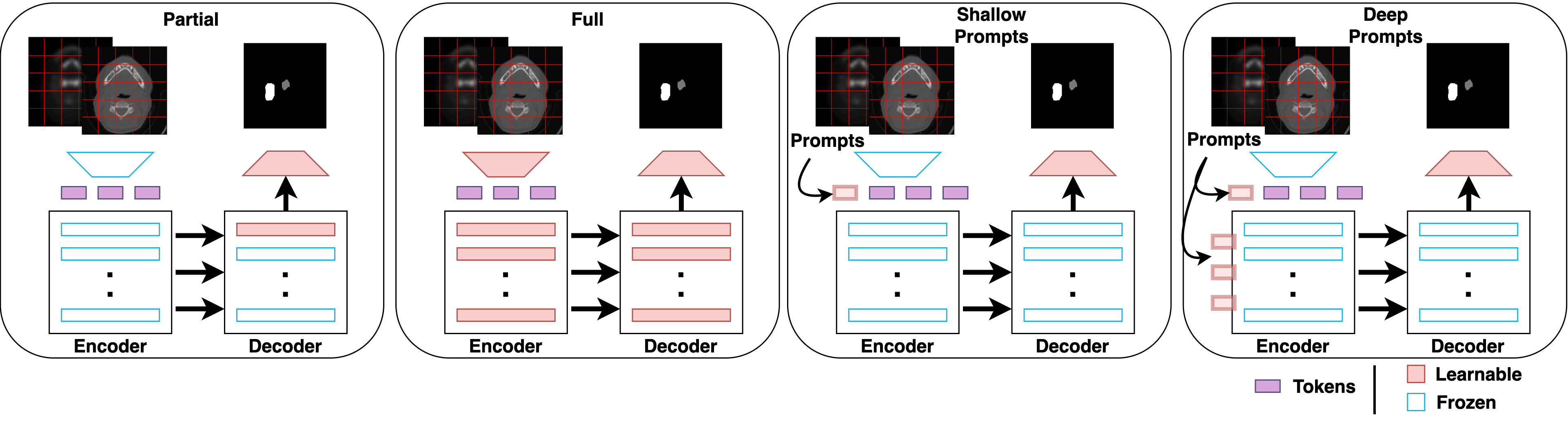}
\caption{Illustrations of the different fine-tuning methods, including partial and full fine-tuning (conventional) as well as shallow and deep prompt-based (proposed).} \label{fig3}
\end{figure}

The dataset for each of the seven centers is first split into train and test sets with a ratio of 70:30, respectively, for a fair comparison. In all experiments, the model is first pre-trained using the six centers' training data and then fine-tuned on the seventh center's training data. We evaluate the performance of the model on (1) the seventh center's test set (new center) and (2) on the six centers' test set (old centers). We compare both metrics for the following fine-tuning techniques as shown in Figure~\ref{fig3}.\\
{\itshape {No fine-tuning:}} 
 In this, the pre-trained model is directly used to infer the test samples without any fine-tuning.\\ 
{\itshape {Partial fine-tuning:}} This technique involves fine-tuning the pre-trained model's last decoder block using the seventh center's training set. \\
{\itshape {Full fine-tuning:}} This technique involves fine-tuning the entire pre-trained model using the seventh center's training set. \\
{\itshape {Shallow prompt fine-tuning:}} This is a variant of prompt-based fine-tuning, where the prompts are introduced only in the input space. Only the prompts and the final convolutional layer are fine-tuned using the seventh center's training set, while the rest of the model is frozen. \\
{\itshape {Deep prompt fine-tuning:}} This technique is similar to shallow prompt fine-tuning; prompts at each level of the transformer layer are introduced. Thus, at each level, there are new trainable prompts to refine. The prompts and the final convolutional layer are fine-tuned using the seventh center's training set.

\subsection{Implementation Details}
We implement all our models using the PyTorch framework and train them on a single NVIDIA Tesla A6000 GPU. The details of the experimental settings for all fine-tuning techniques are listed in Appendix \ref{app1} Table \ref{tab:supp1}. 

All images are aligned to the same 3D orientation (anterior–posterior, right–left, and inferior–superior) during training and testing. The CT/PET scans are concatenated to form a 2-channel input, with their intensity values independently normalized based on their respective means and standard deviations. The training augmentations applied to the CT/PET scans include extracting four random crops of size $96\times96\times96$, with each having an equal probability of being centered around the primary tumor or lymph node voxels and the background voxels. The images are randomly flipped in the $x$, $y$, and $z$ directions, with a probability of 0.2, and are further rotated by 90 degrees in the $x$ and $y$ directions up to 3 times, with a probability of 0.2. These augmentations aim to create more diverse and representative training data, which can help to improve the performance and generalization of deep learning models for medical image analysis tasks. All pre-processing and augmentation details of the data are listed in {Appendix \ref{app1}} 
 Table \ref{tab:supp2}.

\section{Results}

Table~\ref{tab2} presents the results of fine-tuning the pre-trained UNETR and Swin-UNETR on the old and new medical center datasets. We conduct our evaluations using a five-fold cross-validation with a total of 290 experiments. The results of all the folds for all the centers can be found in the Supplementary material. We use Dice score~\cite{bertels2019optimizing} to evaluate the performance of segmentation in our experiments. We can observe that:

\startlandscape
\begin{table}[H]

\tablesize{\footnotesize}
\caption{Aggregated five-fold Dice scores of GTVp and GTVn using different fine-tuning techniques with UNETR and Swin-UNETR.\label{tab2}}
		\newcolumntype{C}{>{\centering\arraybackslash\scriptsize}X}
		\begin{tabularx}{\linewidth}{CCCCCCCCCCCC}
			\toprule
			 \textbf{Model} & \textbf{Fine-Tuning} & \multicolumn{2}{c}{\textbf{None}} & \multicolumn{2}{c}{\textbf{Partial}} & \multicolumn{2}{c}{\textbf{Full}} & \multicolumn{2}{c}{\textbf{Shallow Prompts}} & \multicolumn{2}{c}{\textbf{Deep Prompts}}\\
    
			\midrule
   &\textbf{Center(s)} & \textbf{Old} & \boldmath\textbf{New ($\mu\pm\sigma$)} & \textbf{Old} & \boldmath\textbf{New ($\mu\pm\sigma$)} & \boldmath\textbf{Old} & \boldmath\textbf{New ($\mu\pm\sigma$)} & \textbf{Old} & \boldmath\textbf{New ($\mu\pm\sigma$)} & \textbf{Old} & \boldmath\textbf{New ($\mu\pm\sigma$)}\\
			\midrule
       & CHUP & 0.7869 & 0.6708~$\pm~0.0529$  & 0.7027  & 0.7153~$\pm~0.0496$  & 0.7048 & 0.7298~$\pm~0.0579$  & 0.7507 & 0.7134~$\pm~0.0529$  & 0.7644  & 0.7198~$\pm~0.0565$ \\
       & CHUS & 0.7665 & 0.7699~$\pm~0.0465$ & 0.7574 & 0.7852~$\pm~0.0493$& 0.7574 & 0.7855~$\pm~0.0551$ & 0.7683 & 0.7807~$\pm~0.0513$ & 0.7688 & 0.7831~$\pm~0.0530$\\
       & HGJ & 0.7674 & 0.7877~$\pm~0.0229$ & 0.7479 & 0.7949~$\pm~0.0255$& 0.7439 & 0.7981~$\pm~0.0241$ & 0.7607 &0.7938~$\pm~0.0246$ & 0.7639 & 0.7913~$\pm~0.0259$\\
UNETR  & MDA & 0.7659 & 0.7224~$\pm~0.0231$& 0.7635 & 0.7298~$\pm~0.0209$& 0.7609 & 0.7533~$\pm~0.0245$ & 0.7667 & 0.7339~$\pm~0.0225$ & 0.7657 & 0.7379~$\pm~0.0243$\\
       & CHUV & 0.7704 & 0.7321~$\pm~0.0793$& 0.7622 & 0.7459~$\pm~0.0784$& 0.7665 & 0.7539~$\pm~0.0738$ & 0.7723 & 0.7501~$\pm~0.0759$ & 0.7724 & 0.7531~$\pm~0.0788$\\
       & CHUM & 0.7765 & 0.7714~$\pm~0.0241$ & 0.7584 & 0.7721~$\pm~0.0288$& 0.7623 & 0.7759~$\pm~0.0310$ & 0.7734 & 0.7775~$\pm~0.0266$ & 0.7753 & 0.7799~$\pm~0.0294$\\
       & HMR & 0.7731 & 0.6712~$\pm~0.0489$& 0.7629 & 0.6987~$\pm~0.0580$& 0.7726 & 0.7099~$\pm~0.0496$ & 0.7769 & 0.6992~$\pm~0.0467$ & 0.7760 & 0.7080~$\pm~0.0542$\\
			\midrule
Swin-           &  CHUS & 0.7584 & 0.7695~$\pm~0.0519$ & 0.7569 & 0.7890~$\pm~0.0520$ & 0.7541 & 0.7905~$\pm~0.0458$ & 0.7613 & 0.7797~$\pm~0.0498$ & - & -\\
UNETR           &  CHUM & 0.7763 & 0.7684~$\pm~0.0328$ & 0.7642 & 0.7685~$\pm~0.0370$ & 0.7667 & 0.7706~$\pm~0.0359$ & 0.7719 & 0.7698~$\pm~0.0363$ & - & -\\
                & CHUP & 0.7835 & 0.6609~$\pm~0.0602$ & 0.6960 & 0.7373~$\pm~0.0672$ & 0.7026 & 0.7419~$\pm~0.0592$ & 0.7320 & 0.7136~$\pm~0.0662$ & - & -\\
                &   MDA & 0.7616 & 0.7291~$\pm~0.0257$ & 0.7644 & 0.7413~$\pm~0.0264$ & 0.7541 & 0.7522~$\pm~0.0265$ & 0.7590 & 0.7352~$\pm~0.0263$ & - & -\\  

			\bottomrule
		\end{tabularx}
\end{table}
\finishlandscape

\begin{enumerate}
  \item All the different fine-tuning techniques yield better performance for the new centers than direct inference on the pre-trained models.
  \item Shallow prompt-based fine-tuning achieves a higher or comparable  Dice score on the new-center data, with nearly the same number of learnable parameters as partial fine-tuning (see Table \ref{table3}). However, shallow prompts outperform partial and full fine-tuning techniques on the old-center data for all seven centers. 
  \item Deep prompt-based fine-tuning achieves the same Dice score as full fine-tuning on the new-center data but with significantly fewer learnable parameters. In addition, deep prompt-based fine-tuning outperforms the full fine-tuning on old-center data for all seven centers. Thus, even if the storage of model weights is not a concern, prompt-based fine-tuning is still a promising approach for fine-tuning models as it retains more knowledge related to old centers.
  \item The prompt-based fine-tuning of Swin-UNETR exhibits a similar pattern to that of UNETR. However, the loss in performance on old-center data for the conventional fine-tuning methods is less prominent for some centers compared to that of UNETR. This can be explained by the inductive biases in Swin-UNETR, which employs MSA within local shifted windows and merges patch embeddings at deeper layers. Swin-UNETR requires further optimization with regard to prompt position to further improve its performance.
  
\end{enumerate}

\begin{table}[H] 
\caption{Total number of learnable parameters for different fine-tuning techniques.\label{table3}}
\newcolumntype{C}{>{\centering\arraybackslash}X}
\begin{tabularx}{\textwidth}{CCCCCCC}
\toprule
\textbf{Model}	& \textbf{Fine-Tuning}	& \textbf{None} & \textbf{Partial} & \textbf{Full} & \textbf{Shallow Prompts} & \textbf{Deep Prompts}\\
\midrule
       UNETR &  & -  & 0.025 M & 96 M  & 0.038 M  & 0.15 M \\
\midrule
Swin-UNETR &  & -  & 0.055 M & 62 M  & 0.073 M  & - \\

\bottomrule
\end{tabularx}
\end{table}

        



\section{Discussion}
This work introduces a new method for fine-tuning transformer-based medical segmentation models on new-center data. Our method is more efficient than conventional approaches, requiring fewer parameters at a lower computational cost while achieving the same or better performance on new-center data when compared to conventional methods {(Table~\ref{supp_3})}. 
We show superior performance for prompt-based fine-tuning compared to other techniques, achieving a statistically significant increase in the Dice score for old centers. We note the difference in performance between CHUP and CHUS, which have a similar number of samples but different acquisition machines and origins. CHUP exhibits a larger drop in performance on the old centers than CHUS (nearly 8\% in CHUP vs. 1\% in CHUS for partial and full fine-tuning). This is likely due to the larger dataset distribution shift in CHUP compared to the rest of the centers. However, if shallow- or deep prompt-based fine-tuning is used, the drop is only 2--3\%.
 We perform a Wilcoxon signed-rank test~\cite{Wilcoxon1992} to assess whether the deep prompt-based tuning of medical segmentation models is significantly better than other fine-tuning techniques on old- and new-center data (the null hypothesis $H_0$ states that the  segmentation performance of deep prompt-based fine-tuning is statistically the same as the other techniques. The alternative hypothesis $H_1$ states that the deep prompt-based technique outperforms the other methods). Table~\ref{table4} presents the results of each test; it can be observed that deep prompt-based fine-tuning outperforms full and partial fine-tuning techniques on the old center's data. Similarly, it outperforms the partial prompt- and shallow prompt-based techniques on the new-center data. However, the test fails on the new center's data for full fine-tuning. Thus, we proceed to performing a two-tailed t-test and confirm that the performances of deep prompt-based fine-tuning and full fine-tuning on new-center data are statistically the same ($p$-value $<0.05$).

\begin{table}[H] 
\caption{Wilcoxon signed-rank test on whether deep prompt-based fine-tuning of UNETR performance is better than the other methods.\label{table4}}
\newcolumntype{C}{>{\centering\arraybackslash}X}
\begin{tabularx}{\textwidth}{CCCC}
\toprule
	& \textbf{Partial?}	& \textbf{Full?} & \textbf{Shallow?} \\

\midrule
       Is the performance of deep prompt-based fine-tuned models on {\textbf{old}} 
 centers
        statistically better than   & $\checkmark$ & $\checkmark$ &  \ding{53} \\
        \midrule
        Is the performance of deep prompt-based fine-tuned models on {\textbf{new}} centers
        statistically better than  & $\checkmark$ & \ding{53} &  $\checkmark$ \\

\bottomrule
\end{tabularx}
\end{table}


In our experiments, we observed that the extra learnable prompts at deeper layers in the deep prompt-based fine-tuning improve the performance compared to shallow prompt-based fine-tuning, which only inserts prompts in the input space after the patch embedding layer. {We present the results of ablating different prompt positions and prompt numbers in Tables~\ref{tab:supp-position}--\ref{tab:supp-numPrompts}. Our findings indicate that their specific position does not significantly influence the model's performance when the number of prompts is fixed. However, for a fixed number of prompts distributed across various layers, incorporating prompts into the skip connection layers adversely affects the model's performance, while their exclusion leads to performance improvements, as shown in Table~\ref{tab:supp-skipPrompts}. Furthermore, the results reveal that increasing the number of prompts initially yields improvements in performance. However, there is a threshold beyond which the model tends to become overparameterized, resulting in a degradation of its performance. These results serve as motivation for our choice to position the deep prompts after the skip connection layers in our design}.  This suggests that adding too many prompts in the deeper layers can over-parameterize the model, which may result in overfitting on new-center data. Further studies will be conducted to quantify the effect of the number and position of the prompts.

\section{Conclusions}
We propose a prompt-based fine-tuning framework for the medical image segmentation problem. This method takes advantage of the strength of transformers to handle a variable number of tokens at the input and the deeper layers. We validate our proposed method by training transformer-based segmentation models on head and neck PET/CT scans and compare our results with conventional fine-tuning techniques. Although we were able to show the efficacy of the proposed method on medical image segmentation problems, further investigation is needed to study its scalability to other transformer-based segmentation models in the future. In addition, investigation of prompt-based learning in different tasks, such as classification and prognosis, is needed to assess its efficacy, along with its performance comparison with domain generalization methods.

\vspace{6pt}
\authorcontributions{Conceptualization, N.S., M.R., and M.Y.; methodology, N.S., M.R., and M.Y.; software, N.S., M.R., R.A.M., and M.Y.; validation, N.S., M.R., R.A.M., and M.Y.; formal analysis, N.S. and M.R.; investigation, N.S., M.R., and R.A.M.; resources, M.Y.; data curation, R.A.M.; writing---original draft preparation, N.S. and M.R.; writing---review and editing, N.S., M.R., R.A.M., and M.Y.; visualization, N.S. and R.A.M.; supervision, M.Y.; project administration, N.S. and M.R.; and funding acquisition, M.Y. All of the authors have read and agreed to the published version of the manuscript.}

\funding{M.Y., N.S., M.R., and R.A.M. were funded by MBZUAI research grant (AI8481000001).}

\institutionalreview{Not applicable.
}

\informedconsent{Not applicable.}


\dataavailability{Data used in this study were obtained from the Head and Neck Tumor Segmentation and Outcome Prediction in the PET/CT Images challenge~\cite{oreiller2022head}.

} 




\conflictsofinterest{{The authors declare no conflict of interest. The funders had no role in the design of the study; in the collection, analyses, or interpretation of data; in the writing of the manuscript; or in the decision to publish the results.}}




\abbreviations{Abbreviations}{
The following abbreviations are used in this manuscript:\\

\noindent 
\begin{tabular}{@{}ll}
CT & Computed tomography\\
PET & Positron emission tomography\\
ViT & Vision transformer\\
CNN & Convolutional neural networks\\
MSA & Multi-head self attention\\
\end{tabular}
}

\appendixtitles{no} 
\appendixstart
\appendix
\section[\appendixname~\thesection]{}\label{app1}
\appendix
\appendixtitles{yes} 
\subsection[\appendixname~\thesubsection]{Experimental Settings and Augmentations}

\vspace{-6pt}
\begin{table}[H] 
\caption{Experimental settings for the different fine-tuning techniques. \label{tab:supp1}}
\newcolumntype{C}{>{\centering\arraybackslash}X}
\begin{tabularx}{\textwidth}{CCCC}
\toprule
 \textbf{Hyperparameters} & \textbf{Full} & \textbf{Shallow Prompt} &\textbf{  Deep Prompt}\\
\midrule
Optimizer &  AdamW & SGD & SGD\\
lr   &  1 $\times$ 10$^{-5}$ & 0.05 & 0.05\\
Weight decay &  1 $\times$ 10$^{-3}$ & 0 & 0\\
Learning rate scheduler &  - & cosine decay & cosine decay\\
Total epochs &  100 & 100 & 100\\
Batch size &  3 & 3 & 3\\
\bottomrule
\end{tabularx}
\end{table}


\vspace{-6pt}
\begin{table}[H] 
\caption{Preprocessing and augmentation details.\label{tab:supp2}}
\newcolumntype{C}{>{\centering\arraybackslash}X}
\begin{tabularx}{\textwidth}{CCCC}
\toprule
\textbf{Augmentations}	& \textbf{Axis}	& \textbf{Probability} & \textbf{Size}\\

\midrule
Orientation & PLS  & - & -\\
CT/PET Concatenation &  1 & - & -\\
Normalization &  - & - & -\\
Random crop  &  - & 0.5 & $96\times96\times96$\\
Random flip &  x, y, z & 0.2 & -\\
Rotate by 90 (up to $3\times$) &  x, y & 0.2 & -\\

\bottomrule
\end{tabularx}
\end{table}

\vspace{-6pt}
\begin{table}[H] 
\renewcommand{\aboverulesep}{.1pt}
\renewcommand{\belowrulesep}{.1pt}

\caption{Comparison 
 of training time and GPU consumption between prompt and non-prompt fine-tuning methods.\label{supp_3}}
\newcolumntype{C}{>{\centering\arraybackslash}X}
\begin{tabularx}{\textwidth}{CCC}
\toprule
\textbf{Fine-Tuning}	& \textbf{Runtime (min)}	& \textbf{GPU Consumption (GB)} \\
\midrule
        Partial & 75 & 15.060  \\
\midrule
        Full & 101 & 41.763  \\
\midrule
        Shallow prompt & 76 & 19.275  \\
\midrule
Deep prompt & 78 &  19.361 \\

\bottomrule
\end{tabularx}
\end{table}


\subsection[\appendixname~\thesubsection]{Five-Fold Results per Center}

\vspace{-6pt}
\begin{table}[H] 
\caption{Five-fold results for UNETR on CHUP center.\label{tab:supp-chup_5Fold}}
\newcolumntype{C}{>{\centering\arraybackslash}X}
\begin{tabularx}{\textwidth}{CCCCCC}
\toprule
\textbf{Fold}	& \textbf{No Finetuning}	& \textbf{Partial Finetuning} & \textbf{Full Finetuning} & \textbf{Shallow Prompt} & \textbf{Deep Prompt} \\
\midrule
1 &  0.6112 & 0.6813 & 0.6307 & 0.6302 & 0.6344\\
2 &  0.7442 & 0.7917 & 0.7596 & 0.7519 & 0.7449\\
3 &  0.6399 & 0.6627 & 0.7285 & 0.6910 & 0.6926\\
4 &  0.6919 & 0.7241 & 0.7551 & 0.7417 & 0.7518\\
5 &  0.6663 & 0.7165 & 0.7753 & 0.7523 & 0.7751\\
\midrule
$\mu\pm\sigma$ & 0.6708~$\pm~0.0509$ & 0.7153~$\pm~0.0496$ & 0.7298~$\pm~0.0579$ & 0.7134~$\pm~0.0529$ & 0.7198~$\pm~0.0564$\\

\bottomrule
\end{tabularx}
\end{table}

\vspace{-6pt}
\begin{table}[H] 
\caption{Five-fold results for UNETR on CHUS center.\label{tab:supp-chus_5Fold}}
\newcolumntype{C}{>{\centering\arraybackslash}X}
\begin{tabularx}{\textwidth}{CCCCCC}
\toprule
\textbf{Fold}	& \textbf{No Finetuning}	& \textbf{Partial Finetuning} & \textbf{Full Finetuning} & \textbf{Shallow Prompt} & \textbf{Deep Prompt} \\
\midrule
1 &  0.7861 & 0.8061 & 0.8058 & 0.8035 & 0.8150\\
2 &  0.7825 & 0.7975 & 0.7884 & 0.7886 & 0.7842\\
3 &  0.6875 & 0.6981 & 0.6906 & 0.6903 & 0.6912\\
4 &  0.7947 & 0.8196 & 0.8125 & 0.8103 & 0.8176\\
5 &  0.7987 & 0.8047 & 0.8300 & 0.8107 & 0.8075\\
\midrule
$\mu\pm\sigma$ & 0.7699~$\pm~0.0465$ & 0.7852~$\pm~0.0493$ & 0.7855~$\pm~0.0551$ & 0.7807~$\pm~0.0513$ & 0.7831~$\pm~0.053$\\

\bottomrule
\end{tabularx}
\end{table}

\vspace{-6pt}
\begin{table}[H] 
\caption{Five-fold results for UNETR on CHUM center.\label{tab:supp-chum_5Fold}}
\newcolumntype{C}{>{\centering\arraybackslash}X}
\begin{tabularx}{\textwidth}{CCCCCC}
\toprule
\textbf{Fold}	& \textbf{No Finetuning}	& \textbf{Partial Finetuning} & \textbf{Full Finetuning} & \textbf{Shallow Prompt} & \textbf{Deep Prompt} \\
\midrule
1 &  0.8009 & 0.8017 & 0.8089 & 0.8086 & 0.8133\\
2 &  0.7387 & 0.7343 & 0.7414 & 0.7410 & 0.7392\\
3 &  0.7626 & 0.7559 & 0.7541 & 0.7702 & 0.7733\\
4 &  0.7668 & 0.7690 & 0.7675 & 0.7697 & 0.7629\\
5 &  0.7882 & 0.7995 & 0.8077 & 0.7981 & 0.8048\\
\midrule
$\mu\pm\sigma$ & 0.7714~$\pm~0.0241$ & 0.7721~$\pm~0.0288$ & 0.7759~$\pm~0.0309$ & 0.7775~$\pm~0.0266$ & 0.7799~$\pm~0.0294$\\

\bottomrule
\end{tabularx}
\end{table}

\vspace{-6pt}
\begin{table}[H] 
\caption{Five-fold results for UNETR on CHUV center.\label{tab:supp-chuv_5Fold}}
\newcolumntype{C}{>{\centering\arraybackslash}X}
\begin{tabularx}{\textwidth}{CCCCCC}
\toprule
\textbf{Fold}	& \textbf{No Finetuning}	& \textbf{Partial Finetuning} & \textbf{Full Finetuning} & \textbf{Shallow Prompt} & \textbf{Deep Prompt} \\
\midrule
1 &  0.6368 & 0.6496 & 0.6633 & 0.6555 & 0.6454\\
2 &  0.7516 & 0.7631 & 0.7667 & 0.7609 & 0.7682\\
3 &  0.7832 & 0.8014 & 0.8013 & 0.8091 & 0.8063\\
4 &  0.8243 & 0.8340 & 0.8422 & 0.8339 & 0.8413\\
5 &  0.6645 & 0.6812 & 0.6959 & 0.6910 & 0.7043\\
\midrule
$\mu\pm\sigma$ & 0.7321~$\pm~0.0793$ & 0.7459~$\pm~0.0784$ & 0.7539~$\pm~0.0738$ & 0.7501~$\pm~0.0759$ & 0.7531~$\pm~0.0788$\\

\bottomrule
\end{tabularx}
\end{table}

\vspace{-6pt}
\begin{table}[H] 
\caption{Five-fold results for UNETR on MDA center.\label{tab:supp-mda_5Fold}}
\newcolumntype{C}{>{\centering\arraybackslash}X}
\begin{tabularx}{\textwidth}{CCCCCC}
\toprule
\textbf{Fold}	& \textbf{No Finetuning}	& \textbf{Partial Finetuning} & \textbf{Full Finetuning} & \textbf{Shallow Prompt} & \textbf{Deep Prompt} \\
\midrule
1 &  0.7258 & 0.7284 & 0.7750 & 0.7492 & 0.7571\\
2 &  0.7384 & 0.7457 & 0.7736 & 0.7464 & 0.7538\\
3 &  0.6843 & 0.6970 & 0.7159 & 0.6979 & 0.7000\\
4 &  0.7426 & 0.7500 & 0.7584 & 0.7504 & 0.7515\\
5 &  0.7207 & 0.7279 & 0.7434 & 0.7258 & 0.7271\\
\midrule
$\mu\pm\sigma$ & 0.7224~$\pm~0.0231$ & 0.7298~$\pm~0.0209$ & 0.7533~$\pm~0.0245$ & 0.7339~$\pm~0.0225$ & 0.7379~$\pm~0.0243$\\

\bottomrule
\end{tabularx}
\end{table}

\begin{table}[H] 
\caption{Five-fold results for UNETR on HGJ center.\label{tab:supp-hgj_5Fold}}
\newcolumntype{C}{>{\centering\arraybackslash}X}
\begin{tabularx}{\textwidth}{CCCCCC}
\toprule
\textbf{Fold}	& \textbf{No Finetuning}	& \textbf{Partial Finetuning} & \textbf{Full Finetuning} & \textbf{Shallow Prompt} & \textbf{Deep Prompt} \\
\midrule
1 &  0.7887 & 0.8024 & 0.8062 & 0.7994 & 0.7955\\
2 &  0.7511 & 0.7534 & 0.7622 & 0.7543 & 0.7566\\
3 &  0.8100 & 0.8031 & 0.8075 & 0.8125 & 0.8114\\
4 &  0.8035 & 0.8225 & 0.8262 & 0.8148 & 0.8191\\
5 &  0.7852 & 0.7935 & 0.7883 & 0.7878 & 0.7739\\
\midrule
$\mu\pm\sigma$ & 0.7877~$\pm~0.0229$ & 0.7949~$\pm~0.0255$ & 0.7981~$\pm~0.0241$ & 0.7938~$\pm~0.0246$ & 0.7913~$\pm~0.0259$\\

\bottomrule
\end{tabularx}
\end{table}

\vspace{-6pt}
\begin{table}[H] 
\caption{Five-fold results for UNETR on HMR center.\label{tab:supp-hmr_5Fold}}
\newcolumntype{C}{>{\centering\arraybackslash}X}
\begin{tabularx}{\textwidth}{CCCCCC}
\toprule
\textbf{Fold}	& \textbf{No Finetuning}	& \textbf{Partial Finetuning} & \textbf{Full Finetuning} & \textbf{Shallow Prompt} & \textbf{Deep Prompt} \\
\midrule
1 &  0.6632 & 0.6903 & 0.7132 & 0.7201 & 0.7453\\
2 &  0.7001 & 0.7188 & 0.7265 & 0.7076 & 0.7194\\
3 &  0.6796 & 0.6797 & 0.6933 & 0.6829 & 0.6854\\
4 &  0.5926 & 0.6229 & 0.6400 & 0.6299 & 0.6264\\
5 &  0.7203 & 0.7817 & 0.7762 & 0.7554 & 0.7632\\
\midrule
$\mu\pm\sigma$ & 0.6712~$\pm~0.0489$ & 0.6987~$\pm~0.0580$ & 0.7098~$\pm~0.0496$ & 0.6992~$\pm~0.0467$ & 0.708~$\pm~0.0542$\\

\bottomrule
\end{tabularx}
\end{table}

\subsection[\appendixname~\thesubsection]{Ablation for Prompt Position and Number of Prompts}

\vspace{-6pt}
\begin{table}[H] 
\renewcommand{\aboverulesep}{.1pt}
\renewcommand{\belowrulesep}{.1pt}

\caption{Effect 
 of changing the position of concatenated prompt on the performance of the model on Fold 1 of CHUP center using UNETR.\label{tab:supp-position}}
\newcolumntype{C}{>{\centering\arraybackslash}X}
\begin{tabularx}{\textwidth}{CCCC}
\toprule
\textbf{Position}	& \textbf{Avg Dice}	& \textbf{P-Tumor} & \textbf{Lymph}  \\
\midrule
shallow & 0.6302 & 0.7778 & 0.4827\\
1 &  0.6307 & 0.7793 & 0.4820 \\
2 &  0.6305 & 0.7775 & 0.4834 \\
3 &  0.6303 & 0.7765 & 0.4840  \\
4 &  0.6306 & 0.7774 & 0.4837 \\
5 &  0.6293	 & 0.7775 & 0.4811  \\
6 &  0.6306 & 0.7792 & 0.4820 \\
7 &  0.6295 & 0.7785 & 0.4806  \\
8 &  0.6303 & 0.7783 & 0.4823 \\
9 &  0.6306 & 0.7789 & 0.4823 \\
10 &  0.6304 & 0.7785 & 0.4822 \\
11 &  0.6304 & 0.7789 & 0.4819 \\
12 &  0.6303 & 0.7786 & 0.4819\\
\bottomrule
\end{tabularx}
\end{table}

\vspace{-6pt}
\begin{table}[H] 
\renewcommand{\aboverulesep}{.1pt}
\renewcommand{\belowrulesep}{.1pt}

\caption{Comparing 
 the model performance on Fold 1 of CHUP center while adding prompts on skip connections vs. no prompts on skip connections.\label{tab:supp-skipPrompts}}
\newcolumntype{C}{>{\centering\arraybackslash}X}
\begin{tabularx}{\textwidth}{CCCC}
\toprule
\textbf{Prompts on Skip Connections}	& \textbf{Avg Dice}	& \textbf{P-Tumor} & \textbf{Lymph}  \\
\midrule
 \ding{53} & 0.6342 & 0.7753 & 0.4931 \\
 $\checkmark$ &  0.6289 & 0.7778 & 0.4810 \\

\bottomrule
\end{tabularx}
\end{table}

\vspace{-6pt}
\begin{table}[H] 
\renewcommand{\aboverulesep}{.1pt}
\renewcommand{\belowrulesep}{.1pt}

\caption{Effect 
 of changing the number of concatenated prompts on the performance of the model on Fold 1 of CHUP center using UNETR.\label{tab:supp-numPrompts}}
\newcolumntype{C}{>{\centering\arraybackslash}X}
\begin{tabularx}{\textwidth}{CCCC}
\toprule
\textbf{Number of Prompts}	& \textbf{Avg Dice}	& \textbf{P-Tumor} & \textbf{Lymph}  \\
\midrule
10 &  0.6292 & 0.7781 & 0.4802 \\
30 &  0.6291 & 0.7766 & 0.4816 \\
50 &  0.6302 & 0.7778 & 0.4827 \\
70 &  0.6307 & 0.7788 & 0.4827 \\
90 &  0.6300 & 0.7774 & 0.4827 \\
100 & 0.6294 & 0.7774 & 0.4815 \\
\bottomrule
\end{tabularx}
\end{table}


\begin{adjustwidth}{-\extralength}{0cm}
\newpage
\reftitle{References}

\PublishersNote{}
\end{adjustwidth}

\begin{thebibliography}{999}

\bibitem[Alalwan \em{et~al.}(2021)Alalwan, Abozeid, ElHabshy, and
  Alzahrani]{alalwan2021efficient}
Alalwan, N.; Abozeid, A.; ElHabshy, A.; Alzahrani, A.
\newblock Efficient 3D Deep Learning Model for Medical Image Semantic
  Segmentation.
\newblock {\em Alex. Eng. J.} {\bf 2021}, {\em
  60},~1231--1239.
\newblock {\url{https://doi.org/https://doi.org/10.1016/j.aej.2020.10.046}}.

\bibitem[Valanarasu \em{et~al.}(2021)Valanarasu, Oza, Hacihaliloglu, and
  Patel]{valanarasu2021medical}
Valanarasu, J.M.J.; Oza, P.; Hacihaliloglu, I.; Patel, V.M.
\newblock Medical Transformer: Gated Axial-Attention for Medical Image
  Segmentation.
\newblock In Proceedings of the Medical Image Computing and Computer Assisted
  Intervention---MICCAI 2021, Strasbourg, France, 27 September--1 October 2021; de~Bruijne, M.; Cattin, P.C.; Cotin, S.; Padoy,
  N.; Speidel, S.; Zheng, Y.; Essert, C., Eds.; Springer International
  Publishing: Cham, Swizterland, 2021; pp. 36--46.

\bibitem[Zhou \em{et~al.}(2021)Zhou, Guo, Zhang, Yu, Wang, and
  Yu]{zhou2021nnformer}
Zhou, H.; Guo, J.; Zhang, Y.; Yu, L.; Wang, L.; Yu, Y.
\newblock nnFormer: Interleaved Transformer for Volumetric Segmentation.
\newblock \emph{arXiv} \textbf{2021}, arXiv:2109.03201

\bibitem[Vaswani \em{et~al.}(2017)Vaswani, Shazeer, Parmar, Uszkoreit, Jones,
  Gomez, Kaiser, and Polosukhin]{vaswani2017attention}
Vaswani, A.; Shazeer, N.; Parmar, N.; Uszkoreit, J.; Jones, L.; Gomez, A.N.;
  Kaiser, L.u.; Polosukhin, I.
\newblock Attention is All You Need. \emph{Adv. Neural Inf. Process. Syst.} {\bf 2017},
\newblock {\em 30}, pp. 5998-6008.


\bibitem[Dosovitskiy \em{et~al.}(2020)Dosovitskiy, Beyer, Kolesnikov,
  Weissenborn, Zhai, Unterthiner, Dehghani, Minderer, Heigold, Gelly,
  Uszkoreit, and Houlsby]{dosovitskiy2020image}
Dosovitskiy, A.; Beyer, L.; Kolesnikov, A.; Weissenborn, D.; Zhai, X.;
  Unterthiner, T.; Dehghani, M.; Minderer, M.; Heigold, G.; Gelly, S.;  et~al.
\newblock An Image is Worth 16x16 Words: Transformers for Image Recognition at
  Scale.
\newblock \emph{arXiv} \textbf{2020}, arXiv:2010.11929.

\bibitem[Hatamizadeh \em{et~al.}(2022)Hatamizadeh, Tang, Nath, Yang, Myronenko,
  Landman, Roth, and Xu]{hatamizadeh2022unetr}
Hatamizadeh, A.; Tang, Y.; Nath, V.; Yang, D.; Myronenko, A.; Landman, B.;
  Roth, H.R.; Xu, D.
\newblock UNETR: Transformers for 3D Medical Image Segmentation.
\newblock In Proceedings of the 2022 IEEE/CVF Winter Conference on Applications
  of Computer Vision (WACV), Waikoloa, HI, USA, 3--8 January 2022; IEEE Computer Society: Los Alamitos, CA, USA,
  2022; pp. 1748--1758.
\newblock {\url{https://doi.org/10.1109/WACV51458.2022.00181}}.

\bibitem[Yan \em{et~al.}(2023)Yan, Liu, Xu, Dong, Li, Shi, Zhang, and
  Dai]{YAN2023109432}
Yan, Q.; Liu, S.; Xu, S.; Dong, C.; Li, Z.; Shi, J.Q.; Zhang, Y.; Dai, D.
\newblock 3D Medical image segmentation using parallel transformers.
\newblock {\em Pattern Recognit.} {\bf 2023}, {\em 138},~109432.
\newblock {\url{https://doi.org/https://doi.org/10.1016/j.patcog.2023.109432}}.

\bibitem[Yu \em{et~al.}(2022)Yu, Wang, Hong, Teku, Wang, and
  Zhang]{medical_transfer_learning}
Yu, X.; Wang, J.; Hong, Q.Q.; Teku, R.; Wang, S.H.; Zhang, Y.D.
\newblock Transfer learning for medical images analyses: A survey.
\newblock {\em Neurocomputing} {\bf 2022}, {\em 489},~230--254.
\newblock {\url{https://doi.org/https://doi.org/10.1016/j.neucom.2021.08.159}}.

\bibitem[Yu \em{et~al.}(2017)Yu, Lin, Meng, Wei, Guo, and Zhao]{transfer_1}
Yu, Y.; Lin, H.; Meng, J.; Wei, X.; Guo, H.; Zhao, Z.
\newblock Deep Transfer Learning for Modality Classification of Medical Images.
\newblock {\em Information} {\bf 2017}, {\em 8},~91.
\newblock {\url{https://doi.org/10.3390/info8030091}}.

\bibitem[Karimi \em{et~al.}(2021)Karimi, Warfield, and
  Gholipour]{Karimi2021-zw}
Karimi, D.; Warfield, S.K.; Gholipour, A.
\newblock Transfer learning in medical image segmentation: New insights from
  analysis of the dynamics of model parameters and learned representations.
\newblock {\em Artif. Intell. Med.} {\bf 2021}, {\em 116},~102078.

\bibitem[Wardi \em{et~al.}(2021)Wardi, Carlile, Holder, Shashikumar, Hayden,
  and Nemati]{WARDI2021395}
Wardi, G.; Carlile, M.; Holder, A.; Shashikumar, S.; Hayden, S.R.; Nemati, S.
\newblock Predicting Progression to Septic Shock in the Emergency Department
  Using an Externally Generalizable Machine-Learning Algorithm.
\newblock {\em Ann. Emerg. Med.} {\bf 2021}, {\em 77},~395--406.
\newblock
  {\url{https://doi.org/https://doi.org/10.1016/j.annemergmed.2020.11.007}}.

\bibitem[Chen \em{et~al.}(2020{\natexlab{a}})Chen, Kornblith, Norouzi, and
  Hinton]{simclr}
Chen, T.; Kornblith, S.; Norouzi, M.; Hinton, G.E.
\newblock A Simple Framework for Contrastive Learning of Visual
  Representations.
\newblock  \emph{arXiv} \textbf{2002},	arXiv:2002.05709

\bibitem[Chen \em{et~al.}(2020{\natexlab{b}})Chen, Fan, Girshick, and He]{moco}
Chen, X.; Fan, H.; Girshick, R.B.; He, K.
\newblock Improved Baselines with Momentum Contrastive Learning. \emph{arXiv} \textbf{2020}, arXiv:2003.04297.

\bibitem[Lin and Heckel(2022)]{lin2022vision}
Lin, K.; Heckel, R.
\newblock Vision Transformers Enable Fast and Robust Accelerated {MRI}.
\newblock In Proceedings of the 5th International Conference
  on Medical Imaging with Deep Learning, Zurich, Switzerland, 6--8 July 2022; Konukoglu, E., Menze, B.,
  Venkataraman, A., Baumgartner, C., Dou, Q., Albarqouni, S., Eds.;  2022;
  Volume 172,  pp. 774--795.


\bibitem[Glocker \em{et~al.}(2019)Glocker, Robinson, Castro, Dou, and
  Konukoglu]{glocker2019}
Glocker, B.; Robinson, R.; Castro, D.C.; Dou, Q.; Konukoglu, E.
\newblock Machine Learning with Multi-Site Imaging Data: An Empirical Study on
  the Impact of Scanner Effects. \emph{arXiv} \textbf{2019}, arXiv:1910.04597.
\newblock {\url{https://doi.org/10.48550/ARXIV.1910.04597}}.

\bibitem[Ma \em{et~al.}(2018)Ma, Zhang, Zanetti, Shen, Satterthwaite, Wolf,
  Gur, Fan, Hu, Busatto, and Davatzikos]{ma2018}
Ma, Q.; Zhang, T.; Zanetti, M.V.; Shen, H.; Satterthwaite, T.D.; Wolf, D.H.;
  Gur, R.E.; Fan, Y.; Hu, D.; Busatto, G.F.;  et~al.
\newblock Classification of multi-site MR images in the presence of
  heterogeneity using multi-task learning.
\newblock {\em Neuroimage Clin.} {\bf 2018}, {\em 19},~476--486.
\newblock {\url{https://doi.org/https://doi.org/10.1016/j.nicl.2018.04.037}}.

\bibitem[Barone \em{et~al.}(2017)Barone, Haddow, Germann, and
  Sennrich]{barone2017}
Barone, A.V.M.; Haddow, B.; Germann, U.; Sennrich, R.
\newblock Regularization techniques for fine-tuning in neural machine
  translation. \emph{arXiv} \emph{2017}, arXiv:1707.09920.

\bibitem[Kumar \em{et~al.}(2022)Kumar, Raghunathan, Jones, Ma, and
  Liang]{kumar2022}
Kumar, A.; Raghunathan, A.; Jones, R.; Ma, T.; Liang, P.
\newblock Fine-Tuning can Distort Pretrained Features and Underperform
  Out-of-Distribution. \emph{arXiv} \textbf{2022},  arXiv:2202.10054.
\newblock {\url{https://doi.org/10.48550/ARXIV.2202.10054}}.

\bibitem[Lester \em{et~al.}(2021)Lester, Al-Rfou, and
  Constant]{lester2021power}
Lester, B.; Al-Rfou, R.; Constant, N.
\newblock The power of scale for parameter-efficient prompt tuning.
\newblock {\em arXiv} {\bf 2021}, arXiv:2104.08691.

\bibitem[Li and Liang(2021)]{li2021prefix}
Li, X.L.; Liang, P.
\newblock Prefix-tuning: Optimizing continuous prompts for generation.
\newblock {\em arXiv} {\bf 2021},  arXiv:2101.00190.

\bibitem[Jia \em{et~al.}(2022)Jia, Tang, Chen, Cardie, Belongie, Hariharan, and
  Lim]{jia2022visual}
Jia, M.; Tang, L.; Chen, B.C.; Cardie, C.; Belongie, S.; Hariharan, B.; Lim,
  S.N.
\newblock Visual prompt tuning.
\newblock In Proceedings of the European Conference on Computer Vision, Tel Aviv, Israel,
   23--27 October  2022, pp. 709--727.

\bibitem[Hatamizadeh \em{et~al.}(2022)Hatamizadeh, Nath, Tang, Yang, Roth, and
  Xu]{hatamizadeh2022swin}
Hatamizadeh, A.; Nath, V.; Tang, Y.; Yang, D.; Roth, H.R.; Xu, D.
\newblock Swin UNETR: Swin Transformers for Semantic Segmentation of Brain
  Tumors in MRI Images.
\newblock In Proceedings of the Brainlesion: Glioma, Multiple Sclerosis, Stroke
  and Traumatic Brain Injuries, Granada, Spain, 16 September 2018; Crimi, A.; Bakas, S., Eds.; Springer
  International Publishing: Cham, Switzerland, 2022; pp.~272--284.

\bibitem[Oreiller \em{et~al.}(2022)Oreiller, Andrearczyk, Jreige, Boughdad,
  Elhalawani, Castelli, Vallières, Zhu, Xie, Peng, Iantsen, Hatt, Yuan, Ma,
  Yang, Rao, Pai, Ghimire, Feng, Naser, Fuller, Yousefirizi, Rahmim, Chen,
  Wang, Prior, and Depeursinge]{oreiller2022head}
Oreiller, V.; Andrearczyk, V.; Jreige, M.; Boughdad, S.; Elhalawani, H.;
  Castelli, J.; Vallières, M.; Zhu, S.; Xie, J.; Peng, Y.;  et~al.
\newblock Head and neck tumor segmentation in PET/CT: The HECKTOR challenge.
\newblock {\em Med. Image Anal.} {\bf 2022}, {\em 77},~102336.
\newblock {\url{https://doi.org/https://doi.org/10.1016/j.media.2021.102336}}.


\newpage
\bibitem[Bertels \em{et~al.}(2019)Bertels, Eelbode, Berman, Vandermeulen, Maes,
  Bisschops, and Blaschko]{bertels2019optimizing}
Bertels, J.; Eelbode, T.; Berman, M.; Vandermeulen, D.; Maes, F.; Bisschops,
  R.; Blaschko, M.B.
\newblock Optimizing the Dice Score and Jaccard Index for Medical Image
  Segmentation: Theory {\&} Practice. In Proceedings of the  Medical Image Computing and Computer Assisted Intervention---MICCAI 2019: 22nd International Conference, Shenzhen, China,  13--17 October 2019. 

\bibitem[Wilcoxon(1992)]{Wilcoxon1992}
Wilcoxon, F., Individual Comparisons by Ranking Methods.
\newblock In {\em Breakthroughs in Statistics: Methodology and Distribution};
  Springer: New York, NY, USA, 1992; pp. 196--202.

\end{thebibliography}
\end{document}